# *A Survey on sentiment analysis in Persian: A Comprehensive System Perspective Covering Challenges and Advances in Resources, and Methods*

Zeinab Rajabi[1*] , MohammadReza Valavi[2]

[1] *Department of Electronic and Computer, Malek-Ashtar University of Technology, Tehran, Iran*
Email: *rajabi.ze@gmail.com*
[2] *Department of Electronic and Computer, Malek-Ashtar University of Technology, Tehran, Iran*
Email: valavi@mut.ac.ir

**Abstract**
Intro: Social media has been remarkably grown during the past few years. Nowadays, posting messages on social media websites has become one of the most popular Internet activities. The vast amount of user-generated content has made social media the most extensive data source of public opinion. Sentiment analysis is one of the techniques used to analyze user-generated data. The Persian language has specific features and thereby requires unique methods and models to be adopted for sentiment analysis, which are different from those in English language. Sentiment analysis in each language has specified prerequisites; hence, the direct use of methods, tools, and resources developed for English language in Persian has its limitations.

Method: The main target of this paper is to provide a comprehensive literature survey for state-of-the-art advances in Persian sentiment analysis. In this regard, the present study aims to investigate and compare the previous sentiment analysis studies on Persian texts and describe contributions presented in articles published in the last decade. First, the levels, approaches, and tasks for sentiment analysis are described. Then, a detailed survey of the sentiment analysis methods used for Persian texts is presented, and previous relevant works on Persian Language are discussed. Moreover, we present in this survey the authentic and published standard sentiment analysis resources and advances that have been done for Persian sentiment analysis.

Result: Finally, according to the state-of-the-art development of English sentiment analysis, some issues and challenges not being addressed in Persian texts are listed, and some guidelines and trends are provided for future research on Persian texts. The paper provides information to help new or established researchers in the field as well as industry developers who aim to deploy an operational complete sentiment analysis system.

**Keywords:** Sentiment analysis, Opinion mining, Persian(Farsi) language, Literature review, Low-resource language

## 1- Introduction

Today, the development of social networks and review websites have remarkably enhanced user-generated content. The analysis of user-generated content on the Internet has many benefits for analyzing customer behaviors, understanding customer needs and tendencies, and improving business and e-commerce. In recent years, social networks, blogs, forums and customer review websites have provided individuals with an opportunity to exhibit their opinions on different topics. This has encouraged researchers to study the sentiment analysis field[1].

---

*\*Corresponding author*

Sentiment analysis, also known as opinion mining, refers to a field of study, in which individuals' opinions, sentiments, evaluations, appraisals, attitudes, and emotions towards entities such as products, services, organizations, individuals, issues, events, topics, and their attributes are examined [2, 3]. In this study, the terms 'sentiment analysis' and 'opinion mining' are used interchangeably. Sentiment analysis is associated with information retrieval, natural language processing(NLP), and the data mining of structured/unstructured data. Studies adopting sentiment analysis have had noticeable growth in recent years[4]. Figure 1 shows the number of documents and citations of 'sentiment analysis' filed under WoS (2003–2016)[5]. The growing number of references to papers on sentiment analysis in recent years has highlighted the significance of the topic.

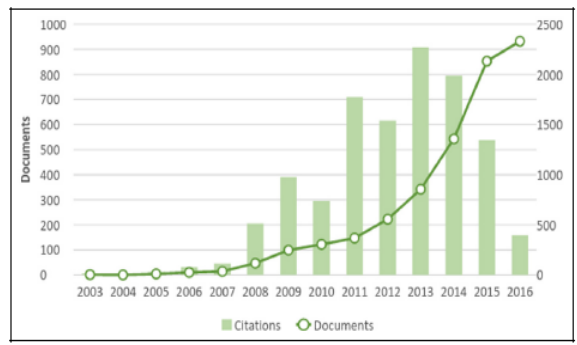

Figure 1 Number of documents and citations of 'sentiment analysis' filed in accordance with the WoS (2003–2016) [5].

Some studies reviewed sentiment analysis methods adopted for texts in English[6-10]. In this regard, studies[11] [12] examined the sentiment analysis methods in Arabic texts; however, sentiment analysis in Persian texts has been disregarded. The Persian language is investigated in different countries. On the other hand, the analysis of Persian texts in international interactions is essential as many such countries need the sentiment analysis results for Persian texts. Some studies (e.g.,[13-16]) highlighted multilingual methods; however, the Persian language was not included. In this addition, many scholars in the field of computer sciences need to know about the position of sentiment analysis in Persian texts to detect what to do, the remaining challenges and gaps in the categorized methods and to investigate the research requirements of the domain. The influence and emotion of human studies in many sciences, including linguistics, psychology, sociology, cognitive sciences, marketing, and communication sciences, has been long considered[17]. Many scholars tend to use the sentiment analysis achievements with a focus on Persian language. Accordingly, the sentiment analysis methods of the Persian texts would contribute to the field. To this end, methodologies applied in Persian texts are surveyed and reviewed in the present study. Furthermore, these methodologies are discussed in the most recent works in terms of sentiment analysis. Finally, some suggestions are put forth to improve further studies in this field.

The first research on the Persian language was performed in 2012; hence, in this paper, we investigated published articles from 2012 to 2020. A total of 34 articles were found during the searches. Table 1 shows the number of reviewed papers over different years. In order to select papers, the queries "sentiment analysis", "opinion mining", "polarity detection", "sentiment classification", "sentiment lexicon" ,"sentiment resource" ,and "sentiment dataset" with "Persian" keyword were submitted to the Google Scholar, science direct, Elsevier, and Springer service. Also, the queries "طبقه بندی قطبیت","تشخیص قطبیت","نظرکاوی","تحلیل سنجمان","تحلیل احساس" were submitted to the Civilica, Magiran, Elmnet, Sid website for published Persian language papers.

Table 1. Number of reviewed papers over different years.

| Year | Number of papers |
|---|---|
| 2012 | 2 |
| 2013 | 3 |
| 2014 | 3 |
| 2015 | 5 |
| 2016 | 5 |
| 2017 | 1 |
| 2018 | 5 |
| 2019 | 5 |
| 2020 | 5 |
| **Total** | **34** |

In summary, the contributions of the present work can be summarized as follows:

- Investigating recently published Persian sentiment papers from different perspectives including problem-solving approaches, contributions of these studies, and evaluation results.
- Investigating recently sentiment analysis methods in Persian texts.
- Investigating recently methods of constructing sentiment lexicon resources for Persian text.
- Reviewing available datasets and classifying them based on important factors.
- Clarifying issues, gaps, and subjects should be noticed in the Persian language regarding the review of all studies in Persian sentiment analysis and also state-of-art results in the English language.

This study is outlined as follows: Section 2 investigates the constraints of sentiment analysis in Persian texts from three perspectives. Section 3 describes different levels of sentiment analysis, Section 4 reviews sentiment analysis approaches, and Section 5 discusses sentiment analysis tasks. Sentiment analysis methodologies in Persian texts are surveyed in Section 6. Methods used to construct sentiment lexicon resources are examined in Section 7. In Section 8, the existing standard Persian datasets are investigated. Section 9 discusses the problems and gaps of sentiment analysis in Persian texts. Finally, some conclusions are provided in Section 10.

## 2- Constraints in Persian sentiment analysis

### 2-1-Preprocessing constraints

Persian alphabet contains 32 letters, and, unlike the English language, words are written from right to left. It has its specific grammar syntax and structure, which is different from other languages. Preprocessing tasks such as sentence tokenizing, word tokenizing, part of speech (POS) tagging, text chunking, lemmatizing, and stemming are used in sentiment analysis approaches. The more accurately the preprocessing is performed in Persian, the more appropriate prerequisites are prepared for the next steps. First, the use of compound words in text has made word tokenizing difficult. Detecting compound words from normal words needs higher accuracy since the compound words such as "پشتیبانی کننده" and "پرداخت کننده" are spaced in between. Second, informal and colloquial words such as "لپ تاپارو", "فک کنم", "واسه" are used many times in opinion texts by users. Third, some words such as "کتاب های" have half-space; however, users do not write half-space when writing their opinions. Therefore, we need more accurate methods for word tokenizing in the Persian language.

Basiri et al.[18] and Asgarian et al.[19] examined the effect of preprocessing on sentiment analysis in Persian texts and approved its effectiveness. Since powerful preprocessing tools are not available for Persian, similar to that used for the English language, some sentiment analysis methods yielding acceptable results in the English texts maybe not achieve the same results in Persian. Moreover, some features of text classification originated from the English texts are not efficient for Persian texts and need further modifications.

### 2-2-Cultural constraints

Individuals' thoughts and subjectivity are significant in polarity identification and sentiment classification. Individuals' perceptions of positive and negative words are different. The culture and characteristics of society affect individuals holding positive or negative opinions. The positive or negative criteria in each society differ from other societies. In this regard, a word may be positive in a society and negative or neutral in another society[20]. A society's perspective toward a sentence may be judged positively; however, another society's perspective may be judged negatively. Sentiment lexicon resources of the English language help to detect polarity even though the resources are constructed publicly and disregard the mentalities of individuals in a society. Each society has special expressions, idioms, and sarcasm intertwined with special sentiment values. The subjectivity and values of society play a significant role in determining polarity. Accordingly, we need resources to match polarity assigned to entities with the culture of society. English sentiment resources and their translation to the Persian language are in contrast with polarity and are not qualified, especially in the political and social domains.

### 2-3-Standard resource constraints

Sentiment analysis methods need text, sentences, and lexicons labeled with sentiment scores. Data accompanied with sentiment scores play a significant role in supervised learning methods. Moreover, deep neural network methods, which have recently been performing well, require massive annotated data. Furthermore, hybrid methods integrate sentiment lexicon resources with learning methods to increase accuracy remarkably. Another challenge in this field is that the preparation of annotated data is complicated and costly.

To tackle this issue, some studies focused on transfer learning model such as bilingual (English, Persian) model[21] and structural correspondence learning(SCL) method to domain adaptation[22].

These resources are constructed in English; however, the Persian language still needs to construct sentiment resources. Some researchers of the Persian texts have been concerned with the standard corpus of the user's opinions at sentence-, feature-, and document-levels as well as their polarity labels.

Another concern in analyzing Persian texts is to construct sentiment lexicon. Labeling sentiment lexicon and domain-specific lexicon is troublesome and time-consuming. Moreover, sentiment lexicon labeling is not performed adequately in many sentiment analysis methods, and linguistic patterns, concepts, and even dependency parsing should be labeled to achieve high performance.

## 3- Different levels of sentiment analysis

In general, sentiment analysis has been mainly investigated at three levels[2]:

**Document level:** The task at this level is to determine whether a whole opinion document expresses a positive or negative sentiment. For example, given a product review, the system specifies whether the review expresses an overall positive or negative opinion about a product. This level of analysis assumes that each document expresses opinions on a single entity (e.g., a single product); hence, document-level sentiment classification is not applicable to documents which evaluate or compare multiple entities.

**Sentence level:** The task at this level is performed on the sentences, i.e. to determine whether each sentence expresses a positive, negative, or neutral opinion.

**Entity and aspect level:** Both the document and sentence level analyses do not discover what individuals exactly liked and disliked. Aspect level performs a finer-grained analysis. Aspect level was earlier called *feature level* (*feature-based opinion mining*). Instead of considering language constructs (documents, paragraphs, sentences, clauses or phrases), the aspect level directly goes through the opinion itself. It is based on the idea that an opinion consists of *sentiment* (positive or negative) and a *target* (of opinion). Opinion targets are described by entities and/or their different aspects. Accordingly, the goal at this analysis level is to discover sentiments of entities and/or their aspects. For example, the sentence "*The iPhone's call quality is good, but its battery life is short*" evaluates two aspects, *call quality* and *battery life* of *the iPhones* (entity). The sentiment of the *iPhone's call quality* is positive; however, the sentiment of its *battery life* is negative. The *call quality* and *battery life* of the iPhones are the opinion targets. According to this analysis level, a structured summary of opinions about entities and their aspects can be produced, which turns unstructured text into structured data and can be used for all kinds of qualitative and quantitative analyses.

## 4- Sentiment analysis approaches

Currently, the most sentiment analysis approach can be divided into three general categories[23]: lexicon-based approach, machine learning approach, and concept-level approach. The main factor in the sentiment analysis field is *sentiment lexicon*, which is commonly used for expressing positive or negative sentiments [2] in terms such as good, excellent, terrible, and bad. Lexicon-based approach is based on sentiment lexicon resources and their sentiment score. It determines the polarity of a text based on the total sum of comprised positive or negative sentiment lexicons in a text. A sentiment lexicon resource includes a set of negative and positive scores assigned to corresponding words and phrases. In comparison with machine learning approach, the lexicon-based method is a suitable approach for sentiment analysis, since it needs fewer resources and does not need some annotated datasets.

Furthermore, a sentiment lexicon can also be used in a machine learning approach in order to make sentiment related features. However, a lexicon-based approach suffers from word coverage of sentiment words in a text. Compared with the lexicon-based approach, a machine learning-based approach uses the most common machine learning algorithms to classify sentiment orientation. Machine learning approach is performed in three ways: supervised learning, unsupervised learning, and semi-supervised learning. Among these, the supervised learning approach has achieved successful results. Feature engineering is critical in classification as the other applications of supervised machine learning. Sentiment classification using supervised learning approach usually formulates the classification problem as a positive and negative class. Many studies used the most common standard classifier for sentiment analysis such as Support Vector Machine (SVM), Naive Bayes and Maximum entropy. This approach was first applied in[24] to classify movie reviews into two classes (positive/negative). They used Naïve Bayes, SVM classifiers, applied unigram, and bag of words as feature for sentiment classification. Later on, many other features and learning algorithms were tested by many scholars in other studies. Some features used by the researchers were sentiment words and phrases, terms and their frequency, POS tag, rules of opinions, sentiment shifters, and syntactic dependency[2, 3]. The approach uses big size previously-annotated data. Training and testing data are normally applied for sentiment classification. Recently, deep neural network learning has also been applied to sentiment analysis and achieved promising results[25].

Some studies such as[26] have adopted the unsupervised method. Since sentiment lexicons are often a major factor for sentiment classification, sentiment words and expressions are used for sentiment classification in an unsupervised manner. The study classifies opinions based on fixed syntactic patterns to be used to express opinions[26]. The syntactic patterns are recognized by POS tags. Then, the probability of co-occurrence of syntactic patterns and sentiment lexicons which have clear polarity by PMI measure in training data are computed.

Cambria et al. presented the third category, known as concept-level sentiment analysis[23, 27, 28] [29]. The concept-level sentiment analysis emphasizes on the semantic analysis of texts with regard to ontology and semantic network, thus providing an opportunity to gather sentimental and conceptual information about different opinions. With an emphasis on a large semantic knowledge base, the approach avoids to count keywords and the number of co-occurrences blindly; however, it relies on the implicit features associated with the linguistic concepts. Unlike purely syntactic methods, the concept-level approach is capable of detecting polarity expressed in a delicate manner. For example, the polarity of a sentence, which is obtained implicitly, can be discovered by concept analysis by linking concepts together. They introduced the CLSA (Concept-Level Sentiment Analysis) framework[30] to support concept-level sentiment analysis. When the concept-level approach is integrated with previous approaches, better results are achieved for sentiment classification[31] [32].

## 6- A review of sentiment analysis methods in Persian texts

Different languages have been spoken around the world, and each language has own specific characteristics. In general, natural language processing methods, particularly sentiment analysis methods, should consider the characteristics of a language. Study[5] mentions languages, to which sentiment analysis research mainly contributes. English, Spanish, Turkish and Chinese are at the top; however, Persian was not included. Persian needs new modified methods because it has specific characteristics and structures. A method for special task of sentiment analysis achieves different results for different languages. The sentiment analysis task was described in the previous section.

Figure 2 shows the sentiment analysis tasks and their adopted methods. *Blue boxes* show the tasks (namely sentiment classification (polarity detection) and the construction of sentiment lexicon resources), on which Persian sentiment analysis studies have been conducted. *White boxes* refer to tasks for which no relevant article was found after reviewing all articles in Persian texts. Accordingly, future studies of Persian texts should be of concern. These include subjectivity detection, sentiment shifters, and fake review detection. *Yellow boxes* refer to aspect/feature extraction, and sarcasm detection for which there are only a few studies, as described in Section 5. *Orange boxes* show methods adopted for two fundamental tasks of polarity detection and the construction of sentiment lexicon resources in the English texts, which are almost applied in Persian texts. As presented in Figure 2, the concept-level sentiment analysis among methods of polarity detection has less been considered (*green* boxes).

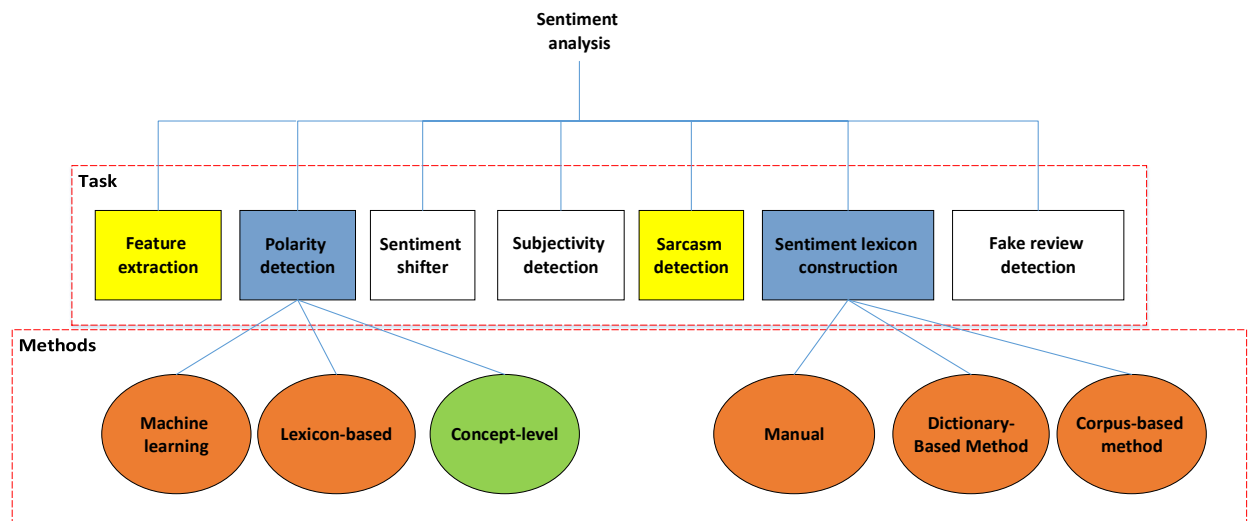

**Figure 2 The categories of sentiment analysis tasks and their methods**

*Blue: Sentiment analysis tasks researched in Persian texts.*

*White: Sentiment analysis tasks with no relevant articles in Persian texts after reviewing various resources.*

*Yellow: Sentiment analysis tasks, for which a few research studies have been found in Persian.*

*Orange: Methods applied for two the tasks of detecting polarity and creating sentiment lexicon in Persian texts.*

*Green: Methods have less been considered in research on Persian texts.*

In this paper, we investigated two tasks of polarity detection and the construction of sentiment lexicon resource as sentiment analysis studies on Persian texts have been mainly focused on these two main tasks. The rest of the tasks are gaps in the Persian language, implying issues for future research.

Shams et al.[33] performed the first sentiment analysis research on Persian texts in 2012. They created the initial seed of sentiment words, called PersianClues, and used the automatic translation of English sentiment lexicon, called Subjectivity Clues, with help of the English to Persian dictionary. They prepared the initial set for the next steps by eliminating possible errors and modifying the sentiment words. Then the LDA-based sentiment analysis method, called LDASA, was presented. This unsupervised sentiment classification method is performed using the initial seed set based on the concerned topic. It determined polarity based on topic and used LDA method to determine topic by weighting words. To evaluate initial sentiment lexicons, the seeds were set from 80 to 160 with 10 intervals and reached 92% to 100% accuracy. Finally, the SVM algorithm was selected to evaluate the whole model, in which the initial seeds were used as features. Accuracy of about 78% was achieved on average in the three domains of hotel, mobile, and camera.

Bagheri et al. conducted two studies[34, 35] on Persian sentiment analysis. They clarified characteristics of the Persian language, including various suffixes of verbs, various plural suffixes, and informal or colloquial words. Next, Modified Mutual Information (MMI) were proposed as appropriate features in Persian text, which improved Mutual Information (MI) measure. MMI was compared to MI and Term Frequency Variance(TFV), and the findings indicated that MMI as features with Naïve Bayesian classifier was better for Persian texts. Moreover, a dataset of 829 user's opinions about mobile products was used for the experimental study.

Vaziripour et al.[36] analyzed the trend of sentiment changes for Iran's nuclear negotiation time. They collected tweets in the political domain during time periods. To identify subtopics, Latent Dirichlet Allocation (LDA) method was applied for four weeks. The study would have some contribution to the field as it was the first study on the Persian language considering the trend of changing sentiment over time and it also performed sentiment analysis dynamically. 274032 tweets from 17844 844 unique users were gathered and annotated. The study reached 70% accuracy by the SVM classifier with Brown cluster as feature.

Amiri et al.[37] used sentiment lexicon-based method for sentiment analysis of Persian texts. They extracted 7179 words, adjective, and expressions from Persian online resources. They included formal, informal, standard, and obsolete words and multi-word expressions, which have an effect on the polarity of texts. The words were annotated via social networks by volunteers having different levels of education, age groups and sectors corresponding to the Persian society. In some cases, when there was no agreement on the polarity label, either the concerned label is discussed or it is determined manually or is labeled as neutral. Finally, sentiment lexicon resources were used for sentiment analysis process and 69% accuracy rate was obtained.

Alimardani and Aghaie [38] [39] proposed hybrid methods (supervised and lexicon-based) for sentiment classification. The study created sentiment lexicon using SentiWordNet in English and defined relationships in FarsNet (Persian WordNet). Three classifier Naïve Baysian, SVM, and logistic regression were applied for improved features by created resource. Present-Absent, TF, and TF-IDF as features are considered and, each feature is multiplied by sentiment value gained from the created resource. In this regard, sentiment lexicon was used as a weighted factor. Then various experimental studies were performed under different conditions, and 85.5% accuracy by SVM classifier with TF-IDF feature was gained in the best conditions. Furthermore, the study investigated different experiments for the influence of the number of positive/negative samples on the accuracy of the model, explained detail of results.

Sadidpour et al.[40] proposed a linguistic-based model to extract the word adjacency patterns to determine the review polarity. The study gathered data by a crawler from the political news. The 208,000 political news were collected from March 2013 to December 2015 on three news sites. Afterwards, they selected their dataset from gathered data. The dataset included about 14,000 news between 2KB and 10KB. They analyzed political subjects in Persian texts and gained about 90% accuracy for the extracted patterns. The positive of the study was that it was conducted on the data from a political domain.

Asgarian et al.[19] performed a complete study on the Persian language in 2018, which had many advantages than others. They also considered the impact of preprocessing on removing stop words, lemmatizing, converting informal to formal words, and spell checking on sentiment analysis. The experimental results of the study indicated that removing stop words and converting informal to formal words had positive effects on Persian sentiment analysis; however, lemmatizing and spell checking had no remarkable effect on Persian sentiment analysis. Moreover, Persian WordNet, called FerdosNet, was constructed and used for the next steps. They created sentiment lexicon resources based on FerdosNet and produced review dataset by using two methods. The first method, English SentiwordNet, was mapped, and the second method used a semi-supervised learning algorithm based on Hidden Markov Model (HMM). Then they did feature engineering for sentiment classification. To this end, some features, which applied the state-of-the-art sentiment classification methods for the English language, were examined for the Persian language. Comparative features included sentiment lexicon, POS tag, sentiment score of lexicon, numeral vectors in word2Vec, sentiment-specific word embedding, character n-gram, word n-gram, TF-IDF. BayesNet, LibLinear, SMO, KNN, Maximum Entropy, and Random Forest classifiers experimented by the mentioned features, whose experimental results were described. The mentioned study presented a roadmap for classifiers and feature selection for those who need to analyze the sentiment of Persian texts.

Dashtipour et al.[41] presented a hybrid framework for the Persian sentiment analysis. They integrated linguistic rules and deep learning to optimize polarity detection. In their work, some dependency rules were extracted for precise sentiment classification. Moreover, the study uses deep neural network models, including CNN and LSTM to further improve the performance. The presented framework was tested by product and hotel review dataset. The various experimental studies were performed under different conditions, and 86.26% accuracy by dependency-based rules and LSTM classifier was gained in the best conditions.

Ghasemi et al. [21] propose a bilingual deep learning framework to benefit from available training data of English. They deployed bilingual embedding to model sentiment analysis as a transfer learning model which transfers a model from a rich-resource language to low-resource ones.

Bilingual sentiment analysis model trains a model on both Persian and English resources and uses it on Persian. The model is evaluated by two datasets from English and Persian which are in the same domain. For English, the electronic product reviews of the Amazon website are used. For Persian, the electronic product reviews of the Digikala website are used. Then various experimental studies were performed under different conditions, and f-measure=91.82% by LSTM-CNN deep neural network architecture based on VecMap embedding methodology was gained in the best conditions.

Table 1 compares studies on the polarity detection of Persian texts. In this Table, the studies are examined in the following terms:

**a) General polarity detection approach:** As mentioned in Section 4, the general approach includes sentiment lexicon-based approach, supervised and unsupervised machine learning approach, and concept-level approach. We examined to detect which research belongs to which categories. As shown in Table 2, [34-36] used supervised learning, [37] used lexicon-based for sentiment classification, and [38] [39] [41] applied the hybrid approach. Moreover, the unsupervised machine learning approach has not been used for classifying documents or sentences in Persian texts. It is worthy to investigate these approaches and examine the results in future research. In addition, concept-level approach has less been considered in research on Persian texts.

**b) Algorithm:** In this column, we explain algorithms used for classification such as Naïve Bayes, maximum entropy, SVM, and logistic regression.

**c) Using features for classification:** The polarity detection task is often modeled as a classification problem, relying on features extracted from the text to feed a classifier. In this column, we compared features concerned in each research study. Classification methods used various features for classifying, as each feature can achieve different performance expressed by accuracy in Table 2.

Various feature such as bag-of-words, terms and their frequency, present-absent word, POS tag, sentiment shifter, and syntactic dependency or their combination were examined to classify English texts. Some of the most commonly used features in sentiment classification are described below:

- **Term and frequency:** Unigram and n-gram along with the frequency of occurrence are common feature for text classification. In some cases, the position of the word may be considered. Weighted TF-IDF from information retrieval was widely used as well. Similar to traditional text classification, these features are effective for opinion classification.
- **Part-of-speech tag(POS):** The POS tag of each word in the sentence can be important too. Words have different POS tags such as noun, adjective, adverb, and verb which have different behaviors. For example, adjectives are significant indicators for detecting opinion; hence, some researchers consider adjectives as special features. Moreover, we can all apply the other POS tags and n-grams as features.

- **Rules of Opinions:** Disregard fewer sentiment words and phrases, many other phrases or their combinations are exploited for expressing and implying opinions and sentiment.
- **Sentiment shifter:** Sentiment shifters cause reverse sentiment orientation (positive to negative or vice versa). The main sentiment shifter is negation words. For example, in the sentence "I don't like this movie" has "like" as a positive word but "don't" as negation word causes reversing sentiment orientation. In addition, there are some other types of sentiment shifter which do not cause shift sentiment orientation in all occurrences so it should be handled carefully. For example, "not" in "not only … but also" does not reverse sentiment orientation.
- **Syntactic dependency:** This feature based on dependency parsing or dependency tree is produced and researchers tried to examine this feature.
- **Sentiment lexicon:** In many studies, sentiment lexicons with their sentiment value are used as feature.

**Table 2 Comparison of sentiment analysis methods in Persian texts**

| Reference | Approach | Algorithm | Feature for classification | Whether the study creates sentiment lexicon or not and use it | Domain | Dataset | Accuracy | Advantage |
|---|---|---|---|---|---|---|---|---|
| Bagheri et al. [34, 35] | Supervised | Naïve Bayes | MMI | No | Mobile product | 829 comments | Recall = 0.8568 | New features are introduced to classify Persian opinions classification. |
| Vazirpour et al. [36] | Supervised | SVM | Brown cluster | No | Political | 274032 tweets from 17844 unique users | Acc =70% | The trend of changes in polarity over time is extracted. |
| Amiri et al. [37] | Lexicon-based | - | - | Yes, by manual | Public | Online source | Precision = 69% | Sentiment lexicon resource is created manually. |
| Alimardani and Aghaie [38] [39] | Hybrid (Supervised and lexicon-based) | Naïve Bayes, SVM, Regression logistic | Present-absent, TF, TF-IDF (each feature was multiplied by sentiment) | Yes, lexicon-based | Hotel | 1805 negative opinions 4630 positive opinions | Best Acc =85.5% | Various experiments were performed on samples at different sizes. The effect of different classifications with three features were investigated. |
| Sadidpour et al. [40] |  | Rule-based (extract pattern) | - | No | Political | 14000 news | Acc =90% | It focuses on the political domain. |
| Dashtipour et al. [42] | Unsupervised | CNN + autoencoder | Feature engineering by deep neural network architecture | No | Movie product | Persian movie reviews | Best acc =82.86% | It evaluates autoencoder and multilayer perceptron for Persian text. |
| Dashtipour et al. [41] | Concept-level | Linguistic rules, LSTM, CNN | Feature engineering by deep neural network architecture | Yes (In previous study) | Product Hotel | Product[†]: 1500 positive 1500 negative Hotel[‡]: | Best acc =86.26% | It integrates linguistic rules and deep learning |

---

[†]http://www.digikala.com

[‡]http://www.hellokish.com).

| Reference | Approach | Algorithm | Feature for classification | Whether the study creates sentiment lexicon or not and use it | Domain | Dataset | Accuracy | Advantage |
|---|---|---|---|---|---|---|---|---|
| | | | | | | 1800 positive and 1800 negative | | |
| Ghasemi et al. [21] | Document-level | Transfer learning, Bilingual embedding, CNN, LSTM, CNN-LSTM, LSTM-CNN | Feature engineering by deep neural network architecture | No | Electronic product | Persian(Digikala): 23,000 English(Amazon): 1,689,188 | Best F-measure =91.82% (LSTM-CNN) | It uses available training data of English to better bilingual embedding for Persian sentiment analysis. |
| [43] | Aspect-level | Unsupervised language-independent (markov chain) | | No | Electronic product | 1200 electronic reviews | Best acc =79.58% | It requires no labelled training data. It can detect aspect and sentiment simultaneously. |
| Roshanfekr et al.[44] | Document-level | Bi-LSTM, CNN | Feature engineering by deep neural network architecture | No | Electronic product | 200761 reviews | Best f-measure =55.4% (CNN) | It evaluates deep learning methods in the Persian language. |
| Zobeidi et al.[45] | Sentence-level | CNN(feature extraction) + Bi-LSTM(sentiment classification) | Feature engineering by deep neural network architecture | No | Electronic product | 151229 reviews (Digikala) | Best acc= 95% (two class) | It uses character-level and word-level input matrix for feature extraction. Classification is done in two classes and in multi-class. |
| Dastgheib et al.[22] | Sentence-level | SRL(structural correspondence learning) + CNN | | No | Electronic product | SentiPers | Best f-measure =74% | It proposed a hybrid method using transfer learning from one domain to similar ones efficiently. |
| Ashrafi Asli et al.[46] | Document-level | Active Learning strategy (LDA sampling)+CNN | Feature engineering by deep neural network architecture | No | Product | Miras Opinion dataset | Best f-measure =80% | It uses active learning with LDA sampling. |

Saraee and Bagheri[47] investigated and compared appropriate features with Naïve Bayesian classifier in the sentiment classification. Four features of document frequency, term frequency variance, MI, and MMI, were applied for the sentiment classification and comparison. In general, findings indicated that MMI feature reached better f-measure than three other features.

Some studies such as[19] performed feature engineering for sentiment classification. To this end, some features which applied for sentiment classification methods of the English language in the state-of-art Persian language. Comparative features included sentiment lexicon, POS tag, sentiment score of lexicon, character n-gram, word n-gram, TF-IDF, numeral vectors in word2Vec, and sentiment-specific word embedding.

Alimardani and Aghaie[38] applied three classifiers (namely Naïve Baysian, SVM, and logistic regression) to improve features by creating sentiment lexicon resources. Present-Absent, TF, and TF-IDF are considered as features, and, each feature was multiplied by sentiment value gained from the created resource. In this regard, sentiment lexicon was used as a weighted factor and an enriching feature.

In general, in the sentiment analysis of each language, feature engineering process often fails to retrieve rich information of relationships and determine polarity that captured from link between concept of text and concepts in the human mind because of feature sparsity and unawareness of the context. Methods using word embedding attempt to eliminate these gaps[48].

**d) Whether the study creates sentiment lexicon or not and use it:** Some studies rely on sentiment lexicons or use sentiment lexicons along with machine learning methods. Since sentiment lexicon resources have been constructed lately, their number is limited and have not efficient in specific applications. Accordingly, some studies aimed to create their own resources using the scientific methods. The column refers to this issue. We investigate studies that their contribution is constructing sentiment lexicon and then using it for polarity detection in more detail in Section 7 and Table 4.

**e) Domain:** Sentiment analysis performs user opinioned data in various domains. In English analysis methods, some studies proposed it generally; however, some studies focused on the specific domain such as product, political, finance, medical and social. The studies in Persian texts were done in various domains but not all domains, as shown in Table 2. The most popular domains are as follows:

- **Product review domain:** The analysis of users' opinions about products and services is the requirement of an e-commerce website. Business owners plan to improve business by getting customer feedback on products and services. Most published studies in a scientific document related to products and services, and the subject such as movies, hotels, restaurants, mobile, laptops, and software are considered.
- **Political domain:** The analysis of users' opinions is about various political subjects on Twitter and news websites. Since specific political terms and irony sentences are used in the political domain, it has unique requirements.
- **Social domain:** It has a very high diversity in the subject. World studies in social domain have less been published on the Internet, and special organizations and governments activated in this domain for social cognition. Sentiment analysis on the social domain is significant for futurology as it has great potentials to study for future research direction.
- **Medicine domain**: It identifies users' opinions about medical issues, drugs, and side effects of drugs.

One method may achieve acceptable accuracy in one domain; however, it may not be efficient in another domain, so the domain of data is essential to analyze. Some domains need more resources to support sentiment classification. In this regard, for example, Noferesti et al.[49, 50] analyzed users' opinions about the drug in the English texts. They constructed a knowledge base for polar facts for the drug domain by using web resources to support sentiment classification.

Due to the importance of the domain, one of the columns in Table 2 is assigned to the domain[38] [34, 35] [21] [43] [44] [44] [45] [22] [46] focused on the products and services domain, and [36, 40] studied the political domain, and[37, 51] [52] analyzed data generally. So far, no research has been conducted in social, medical, and finance domains of the Persian language. Hence, researchers can contribute to these domains in future research.

**f) Dataset**: Studies evaluate the proposed methods by opinion dataset or corpus. Some cases apply to the gathered and published standard data. In contrast, other cases collected data for their own study. Thus, this column reports data gathering resources, number of records, number of positive/negative data, and the number of training and testing data. It is important to know that the accuracy had been achieved for how much data.

**g) Accuracy:** The accuracy of the method was achieved in the studies mentioned in this column. Some studies used accuracy measure, and some others used precision, recall, and f-score measures. Moreover, the methods presented under different conditions (for example, different datasets) reported different accuracy rates for evaluation as such we selected the best accuracy rate and mentioned it in Table 2.

**h) Advantages:** This column refers to advantages of mentioned study.

As shown in Table 2, the number of studies on the Persian language is small compared to those in English, and each of them has dealt with aspects.

## 5- Sentiment analysis tasks

The key sentiment analysis tasks are as follows[8, 30]:

**Subjectivity detection:** In subjectivity detection, opinionated texts are separated from texts containing no opinion. The task is to automatically categorize text into subjective or opinionated (i.e., positive or negative) versus objective or neutral classes.

Sentiment classification (polarity detection): The most basic and well-known task in sentiment analysis is sentiment classification; hence, the terms sentiment classification and sentiment analysis are used interchangeably in many studies. Sentiment classification is to categorize opinions into two classes(positive/negative) or three classes (positive/negative/ neutral) by adopting supervised, unsupervised, or lexicon-based methods.

**Construction of sentiment lexicon resources:** Another fundamental task of sentiment analysis is constructing resources of sentiment lexicon by a scientific approach. In the past years, many studies focused on producing sentiment lexicon resources. The major role of sentiment lexicon resources is to provide the lexicon-based approach. Moreover, the application of the sentiment lexicon to enhance features in the classification of hybrid approaches is approved.

**Aspect/Feature extraction:** Aspect Extraction aims to extract the aspects/features of a product or a service from a text. Aspects are opinion targets, i.e. the specific features of a product or service liked or disliked by users. For example, screen and battery are two features of a mobile phone. We reviewed three studies focusing on this task. Golpar-Rabooki et al.[53] presented a model for aspect extraction based on the iteration of nouns in a text. They improved their work based on the iteration and dual propagation in their second study[54]. In their third study[55], they used syntactic dependency parsing for aspect extraction and achieved results better than those of the two previous methods.

**Sarcasm detection:** Sarcasm is a sophisticated form of speech act, by using which speakers or writers say or write the opposite of what they mean. In the sentiment analysis context, this term indicates that when one says something positive, he/she actually means negative, and vice versa. Sarcasm is a complicated form to be dealt with. Sarcastic sentences are more frequent in online discussions and commentaries on politics. Some studies have addressed this issue in English, but only one research has recently been done in Persian. Golazizian et al.[56] presented first model and first dataset for sarcasm detection. To this end, first, a neural network model pre-trained on a large unlabeled dataset containing Persian tweets with emoji occurrences to predict emojis. Then, it was fine-tuned on a manually labeled irony dataset to detect sarcasm. A BiLSTM network is employed as the basis of model which is improved by attention mechanism. Additionally, a Persian dataset for irony detection, called MirasIrony, containing 4339 manually-labeled tweets is constructed.

**Sentiment shifters:** In sentiment analysis, negation plays a key role in polarity detection and can shift the sentiment value. It may turn a positive polarity into a negative one and vice versa. Accordingly, it is necessary to recognize and apply negation in any sentiment analysis system. The important shifters are negation clues such as "بی", "نا" and "ضد" in a sentence even though they are not sufficient. Negation handling needs deep semantic understanding of the text and is complicated[57].

## 7- Methods of constructing sentiment lexicon resources for Persian text

One of the tasks in sentiment is constructing sentiment lexicon. Sentiment lexicon with sentiment value support lexicon-based and hybrid methods in each language. Hybrid methods apply sentiment lexicon in reinforcing features in sentiment classification. To this end, they created sentiment lexicon resources such as SentiWordNet, General Inquier, MPQA, SenticNet, Vader, etc. in the English language. It is paramount important to create such resources in other languages and use the benefit of them. In this section, we investigate and compare studies, of which their contribution is to construct sentiment lexicon for the Persian language.

Dehdarbehbahani et al.[51] detected polarity value by constructing a multilingual semantic network from external resource WordNet. Semantic network interconnected Persian graph to English graph and utilized random walk algorithm for determining the polarity of Persian words. They reached MAP(Mean average precision) = 0.658 and Accuracy =0.914 in their evaluation results.

Najafzadeh et al.[52] presented a self-training method as a semi-supervised approach for Persian sentiment analysis. This study was the first research on the Persian language using a semi-supervised approach. The method determines sentiment label by self-constructed lexicon (dynamic and without human expert); therefore, it extracts sentiment attributes automatically. Furthermore, Hidden Markov Model classifier of attributes with rule-based usage was employed for the sentiment analysis process. The advantage of this study are: 1) Extracting sentiment attributes automatically instead of human expert 2) Using some colloquial n-grams as a heuristic approach for the Persian language to cover unexplored Persian language challenges, 3) Using ordered pairs (POS tag, polarity tag) as states of Hidden Markov Model to increase the accuracy of Persian sentiment analysis.

Asgarian et al.[58] constructed Persian sentiment lexicon called HesNegar by using FerdosNet lexicon network and mapping English sentiment lexicon (SentiWordNet). To construct HesNegar, the first constructed FerdosNet by mapping Princeton WordNet and English-Persian dictionary, then they mapped sentiment score in English sentiment lexicon to corresponding semantic group of HesNegar. Persian sentiment lexicons were categorized four parts adjective, noun, verb, and adverbial. They reported precision=89% for produced sentiment lexicon. Moreover, a dataset with 3080 user's opinions is annotated by referees and utilized for evaluation.

Dehkharghani[20] proposed a Persian sentiment lexicon called Sentifars by translation method in 2019. He extracted sentiment lexicon from four English sentiment lexicon resources and their translation. Next, sentiment lexicon annotated manually to prepare data for supervised learning logistic regression method. The logistic regression classifier learned the polarity of the Persian lexicon from the polarity score of equivalent in English sentiment lexicon resource. In this case, the polarity of the equivalent lexicon in the English language was determined as a feature in classification. Accuracy=95.92 is best condition as such four resources and their polarities were considered. created resource is produced publicly and domain-independent.

Sabeti et al.[59] proposed LexiPers sentiment lexicon by semi-supervised with FarsNet ontology, selected seed set and labeled by human expert. Then they extended the seed set by using a semi-supervised method and the PMI algorithm.

In general, sentiment lexicon construction methods are divided into three broad categories: 1) Manual method 2) Dictionary-based method 3) Corpus-based method. These three categories and their advantages and disadvantages are discussed in Table 3.

Table 3 Lexicon creation methods and their advantages and disadvantages

| General method | Advantages | Disadvantages | Persian studies which used method |
|---|---|---|---|
| Manual method | It has high accuracy. | It is costly, time consuming and a certain amount of data can be labeled over a period of time. | [37] [60] [61] |
| Dictionary-based method | Dictionaries are available in any language. | The context of the word is not considered (text around the word). <br><br> Public dictionaries do not have information about a domain and are not suitable for providing sentiment lexicon in a domain. <br><br> Dictionary-based methods use synonymous and antonym semantic relations, while dictionaries are not up to date on these relations. | [19] [20] [55] [33] [38] [39] [51] [58] [59] |
| Corpus-based method | Since corpus is usually in a specific domain, they are successful in constructing sentiment lexicon resources in a specific domain. | The resulting lexicons are highly dependent on the corpus data and the lexicons existing in that corpus. <br><br> It is difficult to prepare a corpus having a high coverage. | [62] <br><br> [52] [63] (Based on users rating score from web resources) |

**1) Manual method:** In this method, human experts were asked to label the sentiment value of lexicons and use a majority vote if the expert did not agree with polarity, the majority votes are accepted. The Kappa criterion is used to evaluate confidence value.

The advantage of manual methods compared to the other two methods is high accuracy. Labeling with manual methods is hard and time consuming and has many costs, which are its disadvantages[64]. Only a certain amount of data can be labeled in a given period of time. In some Persian studies (e.g. [61] and [37]) words have been labeled manually.

**2) Dictionary-Based Method:** The second category is methods labeling sentiment lexicon based on dictionaries such as WordNet and English-Persian dictionaries. This method has two advantages. First, English-Persian dictionaries are available. Second, English sentiment lexicon resources which previously prepared labels of polarity are used easily. Further, this method has three disadvantages. The first one is that the context of the lexicon is not considered. Second, public dictionaries have not aware of domain-specific information and appropriate domain-specific sentiment lexicon. Therefore, dictionary-based methods do not notice the context and domain. Third, dictionary-based methods use semantic relations such as synonym and antonym;

however, the relations of dictionaries do not update regularly. The studies [19] [20] [55] [33] [39] [38] [51] [58] for Persian texts are included in this category.

**3) Corpus-based method:** This method extracts sentiment lexicon from a corpus of users' opinions using an unsupervised and supervised manner. The corpus-based method has less accurate than a manual method; however, cost and time of the corpus-based method were affordable.

Advantage of corpus-based method is that it facilitates construct the domain-specific sentiment lexicon more effectively. Because the data of corpus usually belong to a specific domain, constructing domain-specific sentiment lexicon is successful. This method has two disadvantage: First, the extracted sentiment lexicons depend only on the corpus data and the lexicon occurrence in the corpus. Second, it is difficult to preparing corpus which covers whole sentiment lexicon well. Najafzadeh et al.[52] presented a corpus-based method for constructing sentiment lexicon in Persian texts.

Some studies use mixed methods to take advantages of both categories. For example, a small set of sentiment lexicons as seed set is first labeled manually, and then the remained lexicons are labeled based on dataset in a supervised or unsupervised manner. Akhoundzade and Hashemi Devin [62] extracted new sentiment lexicon by cosine similarity( by Word2Vec) and the minimum edit distance from unlabeled data in unsupervised manner.

Since online reviews have rating scores assigned by their reviewers (e.g., 1-5 stars), the positive and negative classes are determined. Moradi et al.[63] used ratting score as measure to determine positive and negative class. This solution achieves acceptable results in the document-level sentiment analysis, while errors in some cases occur in the sentence level sentiment analysis. This is because reviewers usually express both advantages and disadvantages of the sentences so that positive and negative sentences are written together, and no overall rating score can be assigned to each sentence as a sentiment value.

From another perspective, sentiment lexicon construction methods are divided into three categories: manual, automatic, and semi-automatic. Manual methods labeling lexicons are performed manually by human experts. Automatic methods are performed with no human intervention and are only facilitated by machine learning algorithms. Semi-automatic methods apply machine algorithms; however, they assist human experts to increase the accuracy of the method. Accordingly, semi-automatic labeling, which combines two methods, is of a greater value.

In Table 4, we investigate and compare studies which contributed to the creation of sentiment lexicon for the Persian language, and then apply them for polarity detection. The columns of Table 4 are as follows:

**1) General method:** The general method of sentiment lexicon resource construction, described in the previous section, is mentioned in this column.

**2) Partial method of sentiment lexicon resource construction:** The details of construction method are described in this column.

**3) Domain:** The domain of the evaluation or the scope of constructed sentiment lexicon is discussed in studies or some cases domain of evaluation data. As described in Section 6, data can belong to different domain such as political, medical, social, products, and services. As shown in Table 4, most studies [19] [20] [37] [38] [39] [51] [52] [58] [59] [60] in the Persian language present sentiment lexicon publicly.

**4) Dataset:** The column refers to a dataset or corpus used to extract sentiment lexicon. In some cases, the column refers to the dataset used to evaluate the proposed method.

**5) Accuracy of generated sentiment lexicons:** The column refers to the accuracy of generated sentiment lexicons, in which the study was revealed. Some studies reported accuracy measures, and others used precision, recall, and f-score measures used for evaluation. Moreover, whenever presented methods in some studies report different accuracies under different conditions (for example, different datasets) to evaluate, we selected the best accuracy from all values and reported in Table 4.

**6) Advantage:** The column describes the advantage of the proposed method in the study.

**7) Accuracy after using sentiment lexicons in classification algorithm**: After genesis generated by sentiment lexicons, many studies use them in supervised and unsupervised machine learning algorithms. The column refers to accuracy rate after use.

Table 4 Comparison of sentiment lexicon construction methods in Persian texts

| Reference | General method | Partial method of sentiment lexicon resource construction | Domain | Dataset | Accuracy of generated sentiment lexicons | Advantages | Accuracy after using sentiment lexicons in classification |
|---|---|---|---|---|---|---|---|
| Shams et al. [33] | Unsupervised | It used the automatic translation of English sentiment lexicon + Modified LDA method | Hotel, mobile, camera | 200 positive opinion 200 negative opinion | Acc= 91-100% | It determines polarity based on topic and used Unsupervised method. | SVM acc= 78% |
| Dehdarbeh bahani et al. [51] | Lexicon-based and semi-supervised | Random walk algorithm (It Used links between WordNet and Persian semantic graph) | Public | | MAP= 0.658 Acc= 0.914 | The study combines Persian and English semantic networks and uses it to extract polarity | - |
| Amiri et al. [37] | Manual | 7179 words, adjective, and expressions were labeled manually | Public | Online resources | - | It creates a source of sentiment lexicon manually. | Precision= 69% |
| Alimardani and Aghaie [38, 39] | Dictionay-based | It Created the Persian sentiment lexicon by SentiWordNet | Public | - | - | It combines the weight of sentiment lexicon with other features in the classification algorithm. | Best acc= 85.5% |
| Sabeti et al. [59] | Semi-superwised (It used FarsNet ontology as base) | Sentiment seeds was created by manual and extended by PMI method | Public | | Best Acc= 0.81 Best F-measure = 0.66 | If FarsNet is more complete, the expanded version of the algorithm in this article can be used to determine the polarity of new lexicons. | K-nearest neighbors (KNN) nearest centroid (Rocciho) |
| Hosseini et al. [61] | It labeled sentences by manual method. | - | Digital product | - | - | The resource has high accuracy as it is manual. Since it has been published on the | - |

| Reference | General method | Partial method of sentiment lexicon resource construction | Domain | Dataset | Accuracy of generated sentiment lexicons | Advantages | Accuracy after using sentiment lexicons in classification |
|---|---|---|---|---|---|---|---|
| | | | | | | Internet, it can be used for other studies. | |
| Dashtipour et al. [60] | It labeled sentences by manual method. | - | Public | | - | The study creates a new sentiment lexicon resource, including idioms, and then publishes it | SVM acc= 69.29% Naïve Bayes acc= 63.19% |
| Najafzadeh et al. [52] | Semi-supervised | Hidden Markov Model (It created adaptive sentiment lexicon by semi-supervised method) | Public | 190 train data 299 test data | Acc= 89% | The sentiment lexicon resource is construct by semi-supervised methods with no human assessment. It uses a semi-supervised method, which requires less labeled data. | - |
| Asgarian et al. [58] | | It used FerdosNet semantic words and translated English sentiment lexicon to the Persian. | Public | | Precision= 86% F1-measure= 0.8 | It uses FerdosNet vocabulary network, which has a higher coverage than the other vocabulary networks. | - |
| Asgarian et al. [19] | Semi-supervised | Hidden Markov Model | Public | | Precision= 86% F1-measure= 0.8 | It creates a sentiment lexicon resource and used it in different classifications along with different features. | Avg f-measure =88% |
| Dehkharghani [20] | Supervised | It extracted sentiment lexicon based on English resources and used logistic regression classifier for learning polarity of Persian words. | Public | | Acc= 95.92% | It creates the first resource of sentiment lexicon, which gives each lexicon a three-point rating scale (positive, negative, and objective). | - |
| Akhoundzade and Hashemi Devin [62] | Unsupervised | Sentiment lexicons were extracted by cosine similarity (by word2vec) and the minimum edit distance. | Electronic product | 450,000 Reviews (Digikala website) | - | It extracts sentiment words using unlabeled data and supports colloquial words. | Best F-measure=58% |

## 8- Accessible and standard dataset in the Persian language

### 8-1-1-LexiPers

LexiPers Sentiment Lexicon [59] consists of a subset of FarsNet version 2 lexicons which annotated automatically by positive/negative/neutral labels. In this project, the synonymous set of an adjective, number of 4261 synsets, are annotated manually by a human expert as a seed set. The seed set as a gold standard is also used to develop and test the word labeling system and classify documents. The dictionary was produced by the Natural Language Processing and Intelligent Planning Laboratory of Sharif University, and Guilan NLP Group.

### 8-1-Sentiment lexicon resource

There are some sentiment lexicon resources such as SentiwordNet[65], MPQA[66], General Inquirer[67], SenticNet[27], and VADER[68] to sentiment analysis in the English language. Creating and using sentiment lexicons has many benefits in other languages. Few resources have been created for the Persian language, which are described next. Published and accessible resources with their properties are shown in Table 5. The columns are described as follows:

a) **Name:** Name of corpus or dataset which is renowned.

b) **Lexicon/sentence/document/fact:** The column identifies that dataset consists of a lexicon, sentence, document, or fact with polarity value.

c) **Number:** This column reports the number of words, number of sentences, and number of documents.

d) **Domain:** This column reported the domain of sentiment lexicon or dataset which is constructed based on. As described in Section 6, resources can belong to different domains such as political, medical, social, and products and services. Moreover, some resources constructed for a specific domain are not efficient for other domains. For example, the resource produced in the domain of laptop products does not have the efficiency in medical domain. In addition, the resource of sentiment word is usually dependent on the domain [69, 70] and the polarity of words in one domain is sometimes opposite to the other domain.

e) **Publisher:** This column refers to the institution, laboratory, or authors which published the dataset.

Table 5 Existing and published datasets in Persian language

| Name | Sentence | Number | Domain | Publisher |
|---|---|---|---|---|
| LexiPers | Word | 4261 | Public | Natural Language Processing and Intelligent Planning Laboratory of Sharif University and Guilan NLP Group [59] |
| Persian lexicons with polarity label | Word | 3588 adjectives 4073 verbs 7325 nouns | Public | Lab of Intelligent Information Systems at the University of Tehran [51] |
| PerSent | Lexicon, idiomatic words and expressions | 1500 words 700 idiomatic words and expressions | Public | Dashtipour [60] |
| SentiPers | Formal, informal, or colloquial sentences | 110 | Digital product | Guilan NLP Group [61] |
| Hellokish sentiment dataset | Document | 642 | Hotel | Moradi et al. [63] |
| MirasOpinion | Document | 9386 document (49515 positive 14882 Negative 294718 Neutral) | Product | Ashrafi Asli et al.[46] (Miras-tech group) |

### 8-1-2- Persian lexicons with polarity label

Persian lexicons with polarity label[51] which constructed by the lab of Intelligent Information Systems at the University of Tehran, consist of two datasets: 1) A set extracted from annotated Persian adjective: This is constructed by the Persian adjectives of FarsNet. Each entry is specified as positive/negative/neutral. In this regard, more than 3588 adjectives are extracted and evaluated by four referees. 2) A set of adjectives, verbs, and nouns are extracted from FarsNet. Each word in the set assigns sentiment value by the semi-supervised machine learning method. The value smaller than zero is assigned to a negative word, and the value greater than zero assigned to positive words. In the neutral words do not determine explicitly, and neutral words can be determined by a threshold between positive and negative words. The set consists of 3588 adjectives, 4073 verbs, and 7325 nouns, and all words are extracted version 1 of FarsNet.

### 8-1-4- PreSent

Persian Sentiment Analysis and Opinion Mining Lexicon(PreSent) presents real-valued polarity labels, in the range from -1 to 1, for thousands of Persian words and expressions[60, 71]. PerSent 1.0 is the first version of the lexicon including about 1500 Persian words. PerSent 2.0 is the second version of the lexicon, which includes about 1500 Persian words from the first version plus 700 idiomatic words and expressions. The idiom list has been shown particularly useful for analyzing highly informal texts such as user-contributed contents, e.g., movie reviews or product reviews.

### 8-2-Datasets of user opinions

The datasets or corpora of user opinions are other resources which prepared for sentiment analysis. They consist of sentences or documents with positive/negative polarity. In the Persian language, SentiPers and HelloKish datasets were published, the most important are described in the next section.

### 8-2-1-SentiPers

SentiPers[61] consist of Persian sentences with sentiment value used for sentiment analysis. Regarding their features, it is the first dataset for Persian sentiment analysis. The domain of sentences is digital products. Moreover, the sentences of the dataset are formal, informal, or colloquial sentences, and the dataset contains 1100 annotated sentences. It produced by Guilan NLP Group.

### 8-2-2-HelloKish sentiment dataset

HelloKish sentiment dataset[63] annotated by the author's sense and perspective. Dataset gathered from user comments on HelloKish website. When the study was doing, the number of the second registered comments on the website are 3312, and users entered to specify the rate of customer satisfaction in 642 items by website options.

Comments are divided into two categories: opinions whose percentage of satisfaction is ≤ 30% as negative, and opinions whose percentage of satisfaction is ≥ 70% as positive. Accordingly, 102 opinions were placed in a negative category, and 447 opinions were placed in a positive category. The mean of number of words in each opinion is 109 words.

### 8-2-3-MirasOpinion

MirasOpinion dataset[46], is crawled from the Digikala website, one of the largest e-commerce websites in Iran. 2.5 million comments have been crawled, and after some pre-processing, they reduce its size to one million comments. Then the corpus had been labeled using crowd-sourcing; A telegram bot is used to send the unlabeled data to several users. The bot asks them to label the represented document as positive, negative, or neutral. The dataset contains 93868 annotated documents (Positive Comments=49515/Negative Comments=14882/ Neutral Comments 29471). It produced by Miras-tech group.

### 8-3- Results of the review of published dataset

With the advent of machine learning models in sentiment analysis tasks, having a large amount of training data plays an essential role in achieving accurate models. Creating valid training data, however, is a challenging issue in Persian language. As shown in Table 5, from the six published datasets, three datasets labeled lexicon, two datasets labeled the document sentence, and one other dataset labeled the document. Accordingly, we need to construct more Persian datasets, especially at sentence and document levels. Furthermore, we require training data for different domains such as political, social, medical, and drug.

In the lexicon-based approach, the resource covering the lexicons of text and is comprehensive more reaching a better accuracy. Thus, number of sentiment lexicons is important, and comprehensive resources have high performance. Furthermore, the resources are constructed publicly; however, we need to resource with domain-dependent polarity in many applications.

## 9- Remaining challenges and future developments

The Persian sentiment tasks have witnessed a great progress in recent years. However, they suffer from some shortcomings, which make them far from the real polarity detection. In this section, we describe issues and gaps, and subjects should be noticed in the Persian language regarding the review of all studies in Persian sentiment analysis and also state-of-art results in the English language. The extracted main points are as follows, which can be the start point for future research on Persian sentiment analysis:

**1) Sentiment shifter:** As mentioned in Section 5, shifters cause changing sentiment orientation of sentences. However, negation clues such as "نی","عدم" are important shifters, they are not sufficient and shifter management is complicated. One of the challenges in sentiment analysis is shifter management, which was investigated in the English language. Shifter management in the Persian language is less considered and needs to be investigated.

2) **Improve the performance of tokenization:** More powerful preprocessing tools are available in the Persian language, which promote the accuracy of the sentiment analysis methods. Tokenizer has a profound effect on the sentiment analysis. The tokenizer with acceptable accuracy facilitates the preprocessing of sentiment analysis, especially word embedding for deep neural network methods. Thus, it is suggested to study methods and implementations in the Persian language to reach an acceptable accuracy for the tokenizer.

3) **Lack of tools:** Diverse tools, applications, and packages for English language are produced by developers and published on the Internet. They are constructed based on existing various methods. The Persian language suffers from a lack of tools, applications, and packages. The availability of implementations facilitates develops sentiment analysis research in the Persian language. It is suggested to implement tools by scientific methods in Persian language. Furthermore, some research labs develop tools for the advancement of self-works; however, they do not tend to publish them on the Internet and publicly available because of different reasons. This subject performs parallel works and leads to the decreased growth of Persian sentiment analysis.

**4) Concept parsing**: Concept-level sentiment analysis approach considers concepts and implicit semantics of texts. It focusses on the analysis of a text by ontology and semantic networks, which allow the aggregation of effective and conceptual information. Studies on the Persian language less use the approach; therefore, we recommend using the concept-level approach for Persian texts. While concept parser with acceptable accuracy is essential for the concept-level sentiment analysis, we need to study concept parsing methods for Persian text in future works.

**5) Implicit opinion:** Opinion texts are classified into two categories: explicit and implicit opinion. The explicit opinion uses a sentiment lexicon clearly; however, the implicit opinion does not use a sentiment lexicon in text. The implicit opinion is an objective sentence indicating the opinion[2]. For example, the sentence "The device does not antenna" does not contain sentiment lexicon; however, it expresses fact which is negative. The studies on explicit opinion on English text are remarkable and they reach acceptable accuracy. On the other hand, few studies focus on analyzing implicit opinion. The second category presents specific methods[17] [72]; however, it is far away from a comprehensive solution. Many user's opinions submitted on the website are implicit; however, they are removed from standard datasets, thereby better to include them in sentiment analysis research. To this end, we suggest investigating implicit opinions for the Persian language in future research.

**6) Extending methods that require less labeled data:** Since the Persian language has the constraint of the standard dataset, we suggest paying more attention to methods that require fewer data such as transfer learning, domain adaption model, unsupervised, and semi-supervised methods. Unsupervised methods do not require large numbers of data with sentiment label, and semi-supervised methods require less data with sentiment label.

**7) Context sensitive sentiment lexicon:** Since the polarity of sentiment lexicon is not static and depends on the context, many articles examine this subject in English text[73] [74]. As one of the shortcomings, the issue is not addressed in studies on the Persian language. It is suggested to regard context-based sentiment analysis in Persian texts in future research.

**8) Sentiment knowledge base:** Creating a knowledge base with the sentiment value facilitates the sentiment analysis system; Creating domain-specific knowledge. For example, Polarized facts in drug domain have a significant impact on the accuracy of the analysis results. Noferesti et al.[49] examined the effect of polarized facts on the analysis of users' opinions about drugs in English. Future researchers are suggested to create a knowledge base of polarized facts to deal with the sentiment analysis of Persian texts.

**9) Low-resource:** Sentiment lexicon resources and standard annotated dataset for Persian sentiment analysis have constraints. There is no appropriate resource for various domains yet. As shown in Table 5, the existing resources have been mainly created for public and commercial domains, and political, social, and financial domains have been less considered. Moreover, the created resources have low number of data and has no performance in many applications such as deep learning. To this end, we need to construct sentiment lexicons resources and well-defined dataset in various domains in the Persian language.

**10) Domain-specific sentiment analysis methods:** Public sentiment analysis methods are not efficient for some domains. Accordingly, we require customized methods for different domains such as political, social, medical, and drug. To this end, in addition to domain-specific resources such as annotated datasets and knowledge base, we also need to present domain-specific methods for the Persian language in future works.

**11) Deep neural network learning solution:** Deep neural network learning methods have achieved acceptable results in English[75, 76]. Some studies such as [21] [22] [41] [44] [45] [46] [42] have recently applied deep learning models to solve Persian challenges, but it has great potential to study for future research direction. According to the diverse challenges of Persian language, it is possible to apply the learning methods of deep neural networks for Persian sentiment analysis and take advantage from their findings.

**12) Multimodal sentiment analysis:** Nowadays, the sentiment analysis of textual data has highly contributed to the sentiment analysis research. The sentiment analysis of audio and visual data, known as multimodal sentiment analysis, has recently attracted the attention of the scientific community. Sentiment analysis of audio is ignored in research on Persian texts. The analysis of Persian audio data is different from the English ones and requires further studies in the future.

Some challenges such as feature extraction, sarcasm detection, fake review by opinion text, polarity detection of implicit opinion, and so on are less studied in Persian text. In this regard, future studies should investigate them. Furthermore, many challenges have remained for sentiment analysis in the English language. Some studies performed to resolve these challenges. In this regard, the Persian language deserves to be similar to English language to reach such results.

## 10- Conclusion

Sentiment analysis, as a hot research topic related NLP, social network analysis, and computational cognitive science domains focuses on user opinion about entities. The ideal goal of a sentiment analysis system is to gain an understanding of text to be able to determine polarity of text. In this paper, studies on sentiment analysis and opinion mining in Persian language are reviewed. we presented an overview of different aspects of recent Persian sentiment analysis studies, including approaches, methods, tasks, levels, datasets, research contributions, and evaluations. We reviewed 34 Persian sentiment papers from 2012 to 2020 to investigate recent studies and find new trends according to the state-of-the-art development of sentiment analysis. To this end, some methods adopted in main studies on English texts are examined. Existing methods in English language can be generalized to the Persian language; however, they need to be studies delicately and practically with regard to the characteristics of the Persian language, differences of between languages, and preprocessing facilities. In addition, qualified resources such as sentiment lexicon resource, a knowledge base with polarity, dataset with polarity labels in various domains play a significant role in Persian sentiment analysis and should be equipped with scientific approaches.

Finally, we mentioned the future trends and important challenges of available methods including the issues related to sentiment shifter, lack of tools, concept parsing, implicit opinion, context sensitive sentiment lexicon, sentiment knowledge base, low-resource, domain-specific sentiment analysis methods, multimodal sentiment analysis, deep neural network learning solution, improve the performance of tokenization, extending methods that require less labeled data.

**Compliance with Ethical Standards**

**Conflict of interests** The authors declare that they have no conflict of interest.

**Informed Consent** Informed consent was not required as no human or animals were involved.

**Human and Animal Rights** This article does not contain any studies with human or animal subjects performed by any of the authors.

**Informed Consent** Informed consent was not required as no human or animals were involved.

**Human and Animal Rights** This article does not contain any studies with human or animal subjects performed by any of the authors.